%% file: main.tex
\documentclass[nohyperref]{article}

\usepackage{graphicx}
\usepackage{subcaption} 
\usepackage{booktabs} 
\usepackage{csquotes}
\usepackage{natbib}
\usepackage[margin=3.2cm]{geometry}
\usepackage[charter]{mathdesign}

\usepackage{xcolor}
\usepackage[parfill]{parskip}

\usepackage{hyperref}

\usepackage{amssymb}
\usepackage{mathtools}
\usepackage{amsthm}

\usepackage{multirow} 
\usepackage{soul}
\usepackage{colortbl}

\usepackage[T1]{fontenc}

\input{math_commands.tex}

\input{macros}

\usepackage[capitalize,noabbrev]{cleveref}

\theoremstyle{plain}

\theoremstyle{definition}

\theoremstyle{remark}

\newcommand{\papertitle}{Randomness in ML Defenses Helps Persistent Attackers and Hinders Evaluators}
\title{\bfseries\papertitle}

\usepackage[auth-lg,affil-sl]{authblk}

\author[1]{Keane Lucas}
\author[2]{Matthew Jagielski}
\author[3]{Florian Tram\`{e}r}
\author[1]{Lujo Bauer}
\author[2]{Nicholas Carlini}
\affil[1]{Carnegie Mellon University}
\affil[2]{Google Research}
\affil[3]{ETH Zurich}

\begin{document}



\maketitle
\input{sections/abstract}

\input{sections/intro}
\input{sections/relwork}

\input{sections/replay-attack}
\input{sections/make-determ}

\input{sections/sub-space-grid-sweep}
\input{sections/discussion}
\input{sections/conclusion}





\bibliography{cited}
\bibliographystyle{icml2022}

\newpage
\appendix
\onecolumn
\input{sections/appendix}


\end{document}

%% file: math_commands.tex
\usepackage{amsmath,amsfonts,bm}

\def\figref#1{figure~\ref{#1}}
\def\Figref#1{Figure~\ref{#1}}

\def\secref#1{section~\ref{#1}}
\def\Secref#1{Section~\ref{#1}}

\def\eqref#1{equation~\ref{#1}}

\def\floor#1{\lfloor #1 \rfloor}
\def\1{\bm{1}}

\DeclareMathAlphabet{\mathsfit}{\encodingdefault}{\sfdefault}{m}{sl}
\SetMathAlphabet{\mathsfit}{bold}{\encodingdefault}{\sfdefault}{bx}{n}



%% file: macros.tex
\newif\ifdraft
\draftfalse

\newif\ifpublishable
\publishablefalse

\ifdraft
  \newcommand{\todocolor}[1]{\textcolor{cyan}{#1}}
\else
  \newcommand{\todocolor}[1]{}
\fi
\ifdraft
  \newcommand{\commentcolor}[2]{\textcolor{#1}{#2}}
\else
  \newcommand{\commentcolor}[2]{}
\fi

\newcommand{\keane}[1]{\commentcolor{red}{[[Keane: #1]]}}

\newcommand\todo[1]{\todocolor{[[#1]]}}

\newcommand{\tabref}[1]{\mbox{Table~\ref{#1}}}

\newcommand{\appref}[1]{\mbox{App.~\ref{#1}}}

\newcommand{\ml}[0]{ML}

\newcommand{\gridsweep}[0]{Subspace Grid-sweep}
\newcommand{\diffpure}[0]{$\mathit{DiffPure}$}
\newcommand{\ebmdefense}[0]{\textit{EBM--Def}}
\newcommand{\randomsmoothing}[0]{Randomized Smoothing}
\newcommand{\determsmoothing}[0]{Deterministic Smoothing}
\newcommand{\kwta}[0]{k-Winners-Take-All}

\newcommand{\nagfactor}[0]{Nag Factor}

%% file: sections/abstract.tex
\begin{abstract}
It is becoming increasingly imperative to design robust \ml{} defenses. However, recent work has found that many defenses that initially resist state-of-the-art attacks can be broken by an adaptive adversary. 
In this work we take steps to simplify the design of defenses and
argue that white-box defenses should eschew randomness when possible. 
We begin by illustrating a new issue with the deployment of randomized defenses 
that reduces their security compared to their deterministic counterparts. 
We then provide evidence that making defenses deterministic 
simplifies robustness evaluation,
without reducing the effectiveness of a truly robust defense.\keane{does ``increases the likelihood'' invite a quantitative finding (that we don't provide)?}\keane{changed to `simplifies'}
Finally, we introduce a new defense evaluation framework that leverages a defense's deterministic nature to better evaluate its
adversarial robustness.
\end{abstract}

%% file: sections/intro.tex
\section{Introduction}
\label{sec:intro}

Given the increasing prevalence of machine learning models, their reliability in adversarial environments has gained much attention~\citep{Jia2019CertifiedTopkSmoothing,Cohen2019CertifiedAR,rauber2017foolbox,rauber2017foolboxnative, croce2021robustbench,Dong2019-benchmarkadversarialrobust,grathwohl2020ebmdefense,nie2022DiffPure,Madry17AdvTraining}. The ensuing developments towards more robust \ml{} models led to many proposed defenses~\citep{Cohen2019CertifiedAR,grathwohl2020ebmdefense,nie2022DiffPure} and methods to evaluate these defenses~\citep{athalye2018obfuscatedgradients,rauber2017foolbox,croce2021robustbench,rauber2017foolboxnative}.

Unfortunately, evaluating the robustness of defenses 
has proven challenging~\citep{rauber2017foolbox,rauber2017foolboxnative,croce2021robustbench,Dong2019-benchmarkadversarialrobust}. 
Initially promising defenses are often quickly broken~\citep{Xiao2020kWTA,athalye2018obfuscatedgradients}.
One of the leading factors contributing to the difficulty of evaluating robustness is \emph{randomness}~\citep{Gao2022OnTLimitsStochasticDefense}. Prior published work has stated ``Randomization makes the network much more robust to adversarial images, especially for iterative attacks (both white-box and black box)...''~\citep[section 1]{xie2017mitigating} and ``...given an omnipotent adversary, randomness is one way to construct a decision process that the adversary can not trivially circumvent.''~\citep[section 3]{Raff2019BarrageOR}.
We challenge this view, and ask
\begin{displayquote}
\centering
\mbox{\emph{Does randomness improve adversarial robustness?}}
\end{displayquote}
We find that, contrary to popular belief, randomness is neither necessary nor sufficient
for designing robust machine learning defenses---and can even make defenses worse.
This may seem counter-intuitive:
the most effective empirical defenses rely on randomness~\citep{nie2022DiffPure,Cohen2019CertifiedAR},
and many provably robust defenses are randomized \citep{Cohen2019CertifiedAR,Jia2019CertifiedTopkSmoothing}.
Yet we show that, for all defenses we study, the empirical robustness comes not from the randomness of the defense, but from other effects that can be replicated in a deterministic manner.

Specifically we make the following contributions:
\begin{itemize}
    \item We introduce a new threat model that exposes a vulnerability unique to random defenses: the \nagfactor{}, where a random defense allows an attacker to achieve a higher chance of success than when compared with a deterministic version of the same defense, by repeatedly querying the defense (\secref{sec:replay-attack}).
    \item We show that converting a randomized defense to a deterministic version of
    that same defense does not reduce robustness, 
    illustrating that randomness is not the cause of robustness in these defenses (\secref{sec:makedeterm}).
    \item Finally, we introduce a brute-force
    \gridsweep{} method to better evaluate the robustness of \emph{deterministic} defenses (or deterministic variants of randomized defenses), and verify the correctness of a robustness evaluation (\secref{sec:gridsweep}).
\end{itemize}


%% file: sections/relwork.tex
\section{Related Work}
\label{sec:relwork}

\subsection{\ml{} Defense Evaluation}
\label{sec:relwork:defenseeval}

\paragraph{\ml{} Attacks}
Adversarial \ml{} attacks create \textit{adversarial examples}, or inputs that are minimally perturbed to stay within their original class but cause a targeted \ml{} model to misclassify them. To ensure that these perturbations do not change the classification of the input according to a human, they perturb an input at most some distance $\epsilon$ from the original point, under a given distance metric $L_p$ where $p\in\{0,1,2,\infty\}$. Often used attacks include the Fast Gradient Sign Method (FGSM)~\citep{Gdfllw14ExpAdv}, Projected Gradient Descent (PGD)~\citep{Madry17AdvTraining}, and Boundary Attack~\citep{brendel2018boundaryattack}.

Researchers developed standardized suites of attacks~\citep{rauber2017foolbox,rauber2017foolboxnative,croce2021robustbench,Dong2019-benchmarkadversarialrobust} to more fairly benchmark and compare robustness of defenses. These suites include both black-box and white-box attacks, and measure robustness based on how well an \ml{} defense reduces attack success rates. 

\paragraph{\ml{} Defenses}
Researchers have proposed many \ml{} defenses~\citep{Madry17AdvTraining,Pang2018TowardsRobustDetection,nie2022DiffPure,grathwohl2020ebmdefense,Cohen2019CertifiedAR,Jia2018ComDefendAE,Jia2019CertifiedTopkSmoothing,song2018pixeldefend,xie2017mitigating,Raff2019BarrageOR}. This includes adversarial training, where an \ml{} model is exposed to adversarial examples during training, making it less susceptible to attack~\citep{Madry17AdvTraining}. Other defenses include detection of adversarial examples~\citep{Pang2018TowardsRobustDetection}, input refinement~\citep{nie2022DiffPure,grathwohl2020ebmdefense,Jia2018ComDefendAE,song2018pixeldefend}, and other pre-processing techniques~\citep{xie2017mitigating,Raff2019BarrageOR}.

However, the reduced attack success rate does not provide information on \textit{why} the defense fails or succeeds. Prior work shows that these measures can be misleading, giving good scores to defenses that obfuscate gradients, but do not reduce \ml{} model vulnerability~\citep{athalye2018obfuscatedgradients,Tramer2020OnAdaptiveAttacks,Carlini2017AENotDetected}. Other work shows that initially well-performing defenses are vulnerable to adaptive attacks that specifically target weak components of the defense~\citep{Tramer2020OnAdaptiveAttacks}. Our work take steps towards a more useful and informative \ml{} defense evaluation.

Randomness helps with justifying provable robustness certificates~\citep{Cohen2019CertifiedAR,Jia2019CertifiedTopkSmoothing,Levine2019RobustnessCF} and increases robustness to query-based black-box attacks~\citep{Dong2019-benchmarkadversarialrobust} and 
some white-box attacks~\citep{Athalye2018SynthesizingRA,xie2017mitigating}.

However, given the white-box threat model, randomness in many defenses is unhelpful~\citep{Gao2022OnTLimitsStochasticDefense} or gives a false sense of security~\citep{athalye2018obfuscatedgradients,Athalye2018SynthesizingRA}. As previously mentioned in \secref{sec:intro}, a prior published defense states that ``Randomization makes the network much more robust to adversarial images, especially for iterative attacks (both white-box and black box)...''~\citep[section 1]{xie2017mitigating}. This statement is the interpretation of empirical results that successfully mitigate then state-of-the-art attacks by applying random transformations on the input. However, later work overcame this defense by using the same transformations in the attacks that were used in this defense~\citep{Xie2018ImprovingTO}. Related work codified a generalized version of using transformations to improve the effectiveness of attacks called Expectation Over Transformation (EOT)~\citep{Athalye2018SynthesizingRA}. Initially, researchers attributed the gain in robustness to its randomness, which added noise to iterative attacks. However, as this and other random defenses are overcome~\citep{Gao2022OnTLimitsStochasticDefense}, it is unclear whether the \textit{randomness} of a defense is helpful, or an unnecessary component that complicates defense evaluation.

\subsection{Published Randomized Defenses}
\label{sec:relwork:pubdefense}
Random defenses exhibit robustness both empirically~\citep{nie2022DiffPure,grathwohl2020ebmdefense} and provably~\citep{Cohen2019CertifiedAR,Levine2019RobustnessCF,Jia2019CertifiedTopkSmoothing}. In this work, we create deterministic analogs of three of these defenses to understand if randomness is a necessary or useful component against whitebox attacks. 

\paragraph{Randomized Smoothing}
\label{sec:relwork:randomizedsmoothing}
\randomsmoothing{}~\citep{Cohen2019CertifiedAR} is a randomized \ml{} defense which operates by taking the majority vote of the classification of many randomly corrupted copies of the input through the classifier being defended
(with the option to abstain prediction if the vote does not favor one class enough). This method of prediction amounts to a Monte Carlo estimation of the distribution of classes around the input (within some $L_p$ ball), which can be used to justify a robustness certificate on the prediction's correctness~\citep{Cohen2019CertifiedAR}.

\input{figures/devil-replay-attack}

\paragraph{Energy Based Model Defense}
\label{sec:relwork:energybasedmodel}
This defense \citep{grathwohl2020ebmdefense}, referred to as \ebmdefense{} in this work, leverages the idea that a model can be trained to not only classify points in a distribution, but also give a differentiable score on how likely a given input belongs to the data distribution trained on (i.e., an energy function). Using this observation, \ebmdefense{} first adds randomly sampled noise (i.e., a corruption) to the input, then ``refines'' this input by nudging it in the direction of increasing energy, hopefully eliminating any adversarial perturbations not present in the original data distribution. The defense can also 
aggregate multiple parallel corrupt-refine processes (referred to as ``Markov chains'') using different initial corruptions. As 
the refinement process is deterministic, the only random component of this defense is the initial added corruption.

\paragraph{Diffusion Models for Adversarial Purification}
\label{sec:relwork:diffpure}
This defense, referred to as \diffpure{}, \citep{nie2022DiffPure} is similar to \ebmdefense{} in that inputs are corrupted (via random sampling), then \textit{de-noised} (i.e., refined) to remove adversarial perturbations. However, \diffpure{} uses a pre-trained diffusion model separate from the classifier for this refinement and relies on iteratively solving the reverse stochastic differential equation~\citep{nie2022DiffPure}. In contrast, \ebmdefense{} uses stochastic gradient Langevin dynamics~\citep{Welling11LangevinDynamics} with a specially trained energy-based model for refinement that also functions as the underlying classifier\todo{BUT THE DIFFERENCE IS...}\keane{added content}. However, similar to \ebmdefense{}, the only random component in \diffpure{} is the initial corruption of the input.

%% file: figures/devil-replay-attack.tex
\begin{figure*}[!th]
  \centering
  \begin{subfigure}{\textwidth}
    \makebox[\textwidth][c]{\includegraphics[width=0.75\textwidth]{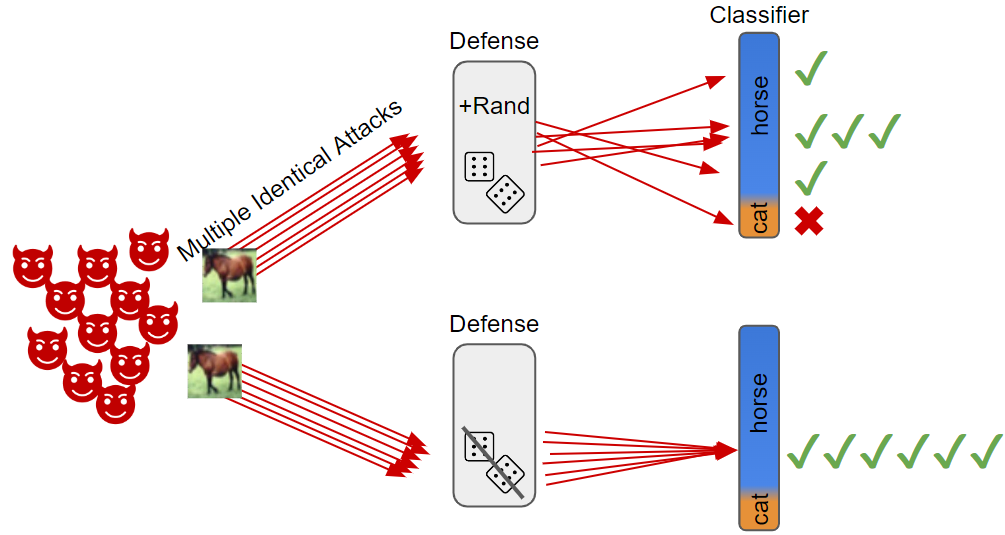}}
  \end{subfigure}

  \caption{\label{fig:devil-replay-attack} The \nagfactor{} threat model assesses the robustness of an \ml{} classifier against an attacker (or group of attackers) that can re-submit their attack many times, increasing the vulnerability of non-deterministic classifiers.}
\end{figure*}

%% file: sections/replay-attack.tex
\section{The \ml{} \nagfactor{}}
\label{sec:replay-attack}
In this section, we consider an attacker whose goal is to achieve at least one misclassification, and has the ability to attempt this misclassification multiple times. This threat would reflect an environment in which querying a model is cheap and not risky, and the reward for fooling or evading the model would be high. For nomenclature, we borrow the term \textit{\nagfactor{}} from research into children's tendency to persistently and repetitively ask for something they want, regardless of their parents' previous refusals, until they are finally given it~\citep{Henry2011NagFactor}.

\paragraph{Threat Model} \todo{fill me in and cut some of the text from above and move it here.}\keane{done}The capability to attempt multiple queries mirror those of an attacker that queries a model many times during an attack to optimize their adversarial input. However, in this case, we assume the attacker only attempts the same input multiple times, with no need to optimize. Because there is no optimization, the attacker only needs to know the black-box hard-label classification for each query. 


\subsection{Multiple Attempts Increase Probability of Success}
\label{sec:replay-attack:methods}
It is easy to see that a non-deterministic classifier (perhaps made non-deterministic by a defense) is always more vulnerable than a deterministic classifier against a persistent and repetitive attacker.

For a classifier $f$ let $p_{(x,y)} = \textbf{Pr}_{r \sim R} [f_\theta(x; r) = y]$
be the probability the stochastic classifier correctly assigns the label $y$ to example $x$ with randomness $r$.
Also, let us denote the accuracy of the determinstic classifier by
$q_{(x,y)} = \textbf{Pr}_{(x,y) \sim X} [f_\theta(x; d) = y]$ with fixed randomness $d$.
To begin, note that $\mathbb{E}_{(x,y)\sim X}[p_{(x,y)}] = \mathbb{E}_{(x,y)\sim X}[q_{(x,y)}]$ on any one random sample
because the fixed randomness $d$, while it does not change, is still randomly selected once.
But now note that if we sample from the random classifier $N$ times, 
choosing a fresh random sample each time,
then the probability that we correctly label the example $x$ every time is exactly $p_{(x,y)}^N$ by independence.
But for the deterministic classifier the probability it returns the correct answer 
remains unchanged, i.e., $q_{(x,y)}$.
Therefore, we are guaranteed that $p_{(x,y)}^N \le q_{(x,y)}$, with the inequality being
a strict inequality any time $p_{(x,y)} < 1$.

\input{figures/robust-acc-vs-replay-plots}

\subsection{Randomness in Deployment Reduces Robustness}
\label{sec:replay-attack:results}
As described in \secref{sec:relwork:randomizedsmoothing}, \randomsmoothing{} makes a prediction by executing 
a Monte Carlo estimation of the distribution of classes around the input by inferring the class of several copies of the input, each with a separate randomly sampled corruption.

To make this process deterministic, we fix the random seed before every prediction and use the same set of corruptions for every inference. We refer to this deterministic version of inference as \textit{\determsmoothing{}}.

We compare the robustness of \determsmoothing{} and \randomsmoothing{} against an attacker willing to simply repeat their inference many times to get at least one misclassification. Mirroring the original \randomsmoothing{} work~\citep{Cohen2019CertifiedAR}, we use CIFAR-10~\citep{cifar} test datapoints for our experiments. This comparison was completed by computing the classification result of 100,000 Gaussian-corrupted copies of each test datapoint using the \textit{base classifier} (i.e., the classifier being defended). As in prior work, this base classifier has been trained with Gaussian-sampled noise-corruptions up to an $L_\infty$ radius of $0.5$.~\citep{Cohen2019CertifiedAR} on the CIFAR-10 training set.

In both \randomsmoothing{} and \determsmoothing{}, the number of inferences $n$ aggregated for a single prediction can be adjusted to control for performance and accuracy requirements. We vary this parameter to observe its affect on robustness. With a setting as low as $n=1$ (which is far below the $n=100$ used in the original work~\citep{Cohen2019CertifiedAR}), \randomsmoothing{}'s prediction is the class inferred from one noise-corrupted inference through the base classifier. This setting predictably leads to a higher variance of predictions given the same input, while higher values of $n$ leads to lower variance predictions. In contrast, \determsmoothing{} with $n=1$ (or $n$ equal to any positive integer), uses the same corruption(s) to corrupt the image for every prediction, so there is no variance in predictions on the same input.

To calculate predictions from \randomsmoothing{} using different values of $n$, we shuffle and group the 100,000 base classifier inferences into groups of size $n$, then calculate each of those groups' mode to get the most commonly inferred class within each group. From these $\floor{\frac{100000}{n}}$ predictions, we then calculate the probability of misclassification for each individual datapoint. In contrast, \determsmoothing{}'s prediction does not change given the same datapoint and value of $n$. 
For this reason, we only calculate one prediction 
for each $(\text{test datapoint},n)$ pair.
Therefore, as \determsmoothing{} either gets its one prediction right or wrong, its probability of correct classification for each test datapoint is calculated as $0\%$ or $100\%$.

Given these probabilities for each datapoint, \figref{fig:robust-acc-vs-replay-attack-plots:diff-corrupt} compares the robust accuracy of both \randomsmoothing{} and \determsmoothing{} over the test dataset using different numbers of corruptions $n$ as we increase the number of trials an attacker is allowed. 
In all cases, \randomsmoothing{} is found to be less robust than \determsmoothing{}, even at the published defense's chosen value of $n=100$~\citep{Cohen2019CertifiedAR}.

As mentioned in \secref{sec:replay-attack:methods}, the probability a classifier (assuming independent trials) $f$ predicts the class of a given datapoint correctly $N$ times in a row is $p^N$ where $p$ is the probability any given prediction on that datapoint is correct. However, as just mentioned, deterministic classifiers correctly classify a datapoint in either $0\%$ or $100\%$ of trials, which means $p^N = p$ for deterministic classifiers. This is why \determsmoothing{} accuracy appears as a horizontal line in \figref{fig:robust-acc-vs-replay-attack-plots:num-trials}, but \randomsmoothing{}'s robust accuracy deteriorates as we increase the number of trials.

Overall, including randomness in an \ml{} classifier (i.e., by using a random \ml{} defense) exposes it to the \nagfactor{}, directly reducing its adversarial robustness.

%% file: figures/robust-acc-vs-replay-plots.tex
\begin{figure*}[!th]
  \centering
  \begin{subfigure}{0.48\textwidth}
    \makebox[\textwidth][c]{\includegraphics[width=\textwidth]{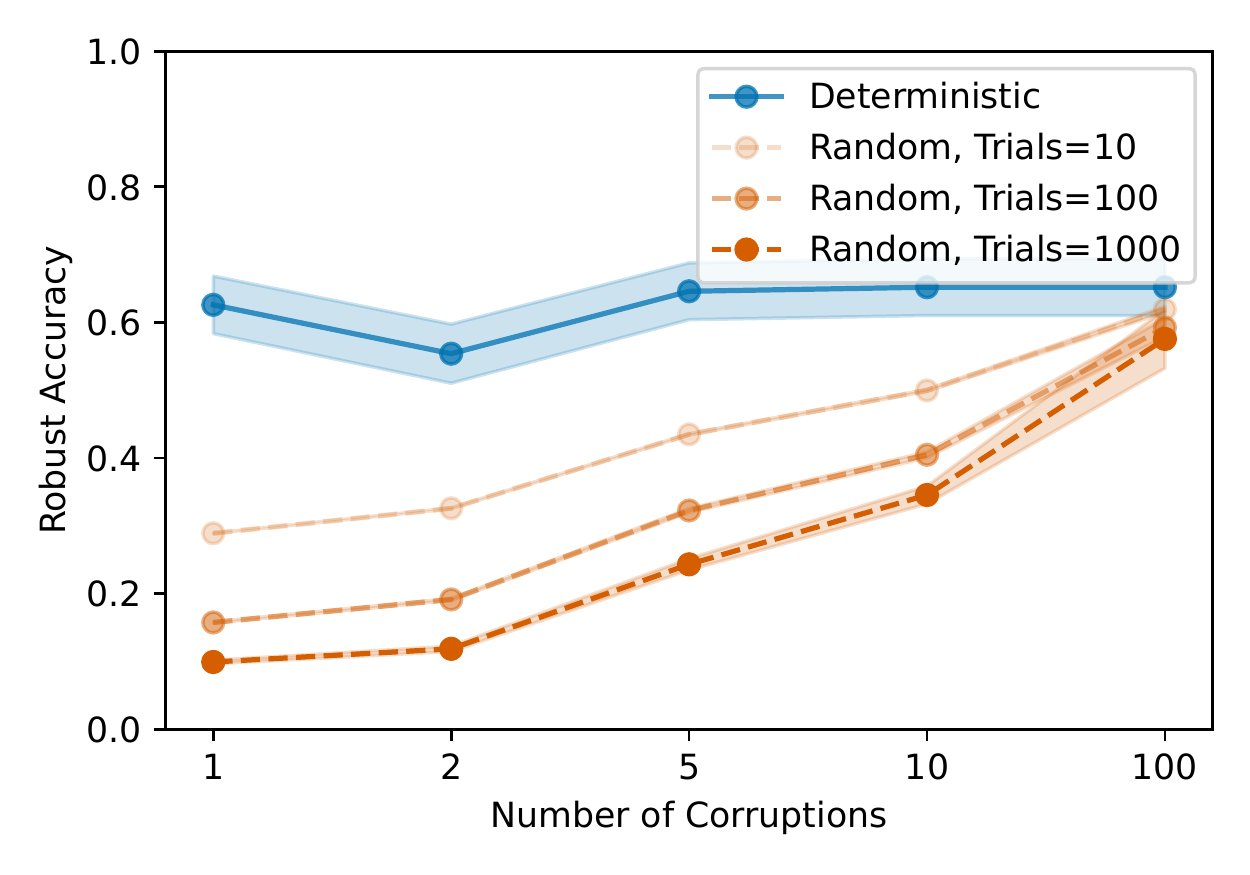}}
    \caption{\label{fig:robust-acc-vs-replay-attack-plots:diff-corrupt} Robust accuracy of randomized smoothing vs deterministic smoothing at varying corruptions. Shaded regions indicate 95\% confidence intervals.}
  \end{subfigure}
  \begin{subfigure}{0.48\textwidth}
    \makebox[\textwidth][c]{\includegraphics[width=\textwidth]{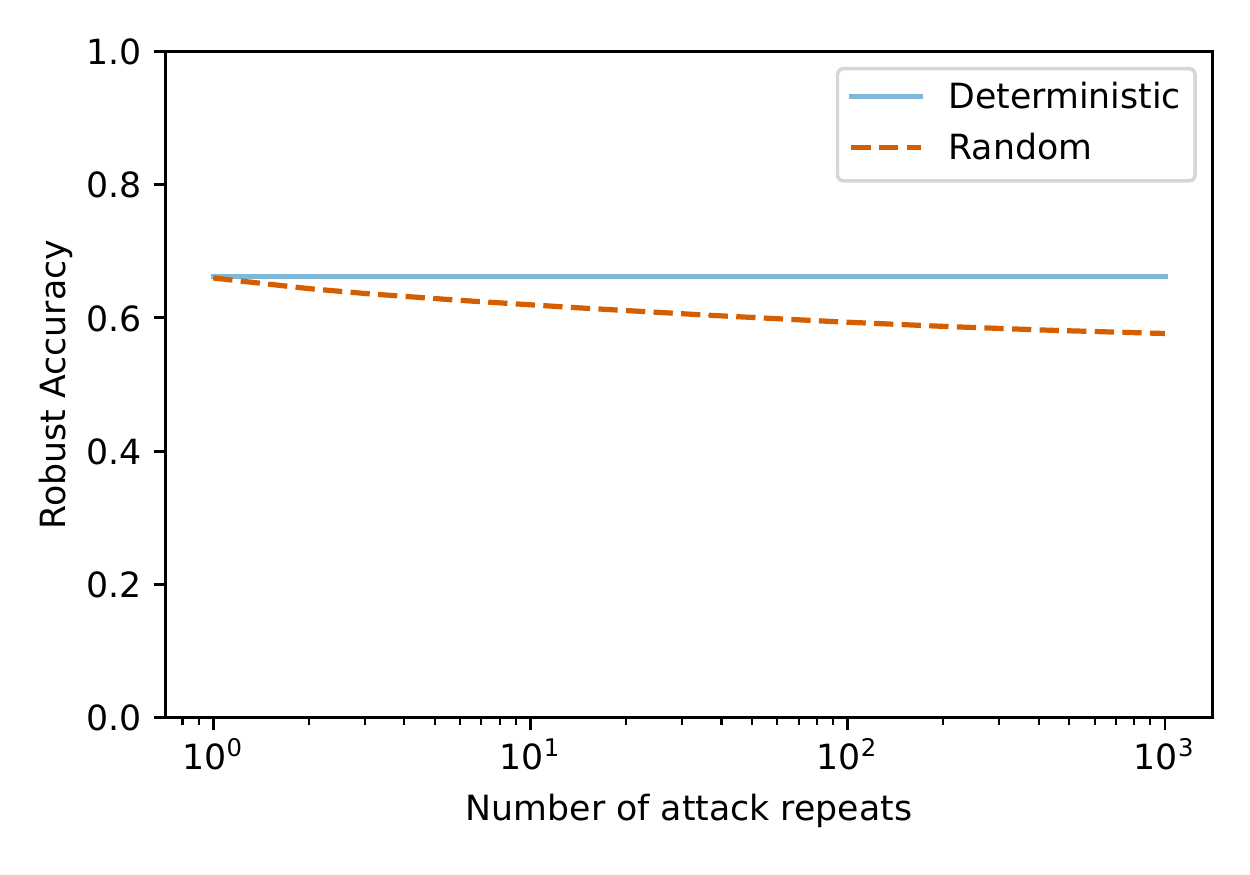}}
    \caption{\label{fig:robust-acc-vs-replay-attack-plots:num-trials} Robust accuracy of randomized smoothing vs deterministic smoothing at 100 corruptions.}
  \end{subfigure}

  \caption{\label{fig:robust-acc-vs-replay-attack-plots} As an attacker repeatedly attempts to evade correct classification, the robust accuracy of a randomized classifier decreases below its deterministic analog.}
\end{figure*}

%% file: sections/make-determ.tex
\section{Deterministic Defenses are Similarly (White-box) Robust}
\label{sec:makedeterm}
The previous section demonstrated that vulnerability can increase when using randomness in deployment. In this section, we evaluate whether or not randomness is useful even in the strongest white-box settings of defenses. 

To do this, we begin by constructing deterministic variants of each of the three randomized defenses introduced earlier (\randomsmoothing{}~\citep{Cohen2019CertifiedAR} (\secref{sec:makedeterm:randomizedsmoothing}), Energy-based Models as a Defense~\citep{grathwohl2020ebmdefense} (\secref{sec:makedeterm:ebmmodels}), and Diffusion Models for Adversarial Purification~\citep{nie2022DiffPure} (\secref{sec:makedeterm:diffpure})). We then compare the robustness of the random and deterministic variants of these defenses by executing Projected Gradient Descent (PGD)~\citep{Madry17AdvTraining} against both of them. 

For consistency, we used each defense's published code and the same implementation of PGD that the original works used to evaluate both variants of the defenses. Specifically, for \randomsmoothing{}, we implemented PGD as described in the original paper~\citep{Cohen2019CertifiedAR}. For \ebmdefense{}, we used foolbox~\citep{rauber2017foolbox,rauber2017foolboxnative} to execute PGD with a binary-search. Finally, we executed PGD from AutoAttack~\citep{croce2021robustbench}, a suite of standardized \ml{} attacks, against \diffpure{}. All attacks were executed, as in the original works, by using test datapoints from the CIFAR-10 dataset~\citep{cifar} as inputs.

Overall, we demonstrate that deterministic defenses have similar or equal empirical robustness to the originally published randomized defenses. As these examples show, robustness exhibited by these defenses emerges from a deterministic characteristic or mechanism and not as the result of any randomness in the defense.

\subsection{Randomized Smoothing}
\label{sec:makedeterm:randomizedsmoothing}
As described in \secref{sec:relwork:randomizedsmoothing} and \secref{sec:replay-attack:results}, \randomsmoothing{} predicts a class by inferring multiple copies of the input through a \textit{base classifier} (the classifier being defended) with added Gaussian-sampled corruptions.

\paragraph{Making this defense deterministic} 
As described in \secref{sec:replay-attack:results}, we fix the random seed before every prediction and use the same set of corruptions for every inference, creating \determsmoothing{}.

\paragraph{Comparing the random and deterministic versions}
\todo{I'm confused why we re-introduce the defense again without much new detail.}\keane{It was intended to be a refresher for the reader, but likely not needed as the previous section is all about it. Commented out.}

\input{figures/robust-acc-vs-num-corruptions}

\todo{Is this a consistent evaluation methodology? If so we shouldn't be putting it here.}\keane{Is your concern that this method could be applicable for other sections? If so, I think this methodology is only applicable for this section (\secref{sec:makedeterm:randomizedsmoothing}) as we only execute PGD against \randomsmoothing{} and \determsmoothing{} here. \Secref{sec:replay-attack} was just repeated (unoptimized) examples.}


\Figref{fig:robust-acc-vs-num-corruptions} shows the robust accuracy of these classifiers as the number of noise-corrupted copies used to make predictions $n$ increases. The robust accuracy of both \randomsmoothing{} and \determsmoothing{} quickly converge to the same value at around $5$ corruptions, which is many fewer than the $100$ corruptions used in the original work~\citep{Cohen2019CertifiedAR}. This indicates that the empirical robustness of \randomsmoothing{} does not come from the randomness of the noise, but the \textit{self-ensembling} effect of aggregating multiple inferences within proximity of the original point.

\subsection{Energy-based Models}
\label{sec:makedeterm:ebmmodels}
Recall from \secref{sec:relwork:energybasedmodel} that the only source of randomness is the random corruption added before the deterministic refinement stage using the trained energy-based model (EBM). 

\paragraph{Making this defense deterministic}
Similar to \randomsmoothing{}, we make the initial added corruptions the same for each inference by setting the random seed to be the same before each prediction.
\todo{Why is there so little here but so much for randomized smothing?}\keane{The \textbf{Making this defense deterministic} paragraphs are currently similar size for \randomsmoothing{} (\secref{sec:makedeterm:randomizedsmoothing}) and this section (it may have been different sizes before and someone else edited?). Or are you referring to \secref{sec:replay-attack:results} giving details about the inference differences between \randomsmoothing{} and \determsmoothing{} to show why \randomsmoothing{} always does worse against a repetitive attacker?}

\input{figures/energy-based-model-converge-plot}

\paragraph{Comparing the random and deterministic versions}
\keane{essentially there's really no difference in the results when making this defense deterministic.}
\todo{I don't understand what this sentence is saying. This is just re-introducing the defense? We should be evaluating in this sentence.}\keane{It is just re-introducing the defense, and mentioning the dataset (CIFAR-10). I moved the dataset mention to following paragraph and commented out this sentence. If we want the first sentence to be eval, should I move the first sentence of the second paragraph up?}


The deterministic and random versions of the defense behave nearly identically. \Figref{fig:ebm-converge-plot} shows the robustness of these versions of \ebmdefense{}, using 1, 2 or 5 ``markov chains'' (i.e., parallel noise-refine processes on copies of the input) to make the prediction given four different thresholds of adversarial example distance from the original test datapoint. These results indicate that any robustness gained by \ebmdefense{} is due to the deterministic refinement, not the randomness of the initial corruptions. For more detailed plots showing a comparison of adversarial example distances that mirrors the original work, see \figref{fig:ebm-dist-plot} in \appref{sec:app:ebmdefense}.

\subsection{Diffusion Models}
\label{sec:makedeterm:diffpure}
Similarly, Diffusion Models for Adversarial Purification~\citep{nie2022DiffPure} uses a similar process of adding randomly sampled noise, followed by a refinement stage (called ``denoising'' or ``purification'' in this work) that uses a trained deterministic diffusion model. 

\paragraph{Making this defense deterministic}
Similar to \determsmoothing{} and \ebmdefense{}, we reset the seed to be the same for the initially added corruptions, ensuring the noise used was the same each time. Furthermore, to create a middle-ground between this deterministic version and the random original, we introduce increasingly more variability by cycling through $k$ different seeds in subsequent inferences, where $k\in\{1,2,3,5,10,50,100\}$.
\todo{We're saying the same thing again without explaining much new. I don't learn anything from this paragraph.}\keane{The first sentence is essentially the same as the other two defense's \textbf{Making this defense deterministic} paragraphs, but I think the cycling through more than one seed is new and not mentioned before. I had liked the look of uniformity of using the same paragraph headers in each defense's section, and in reality making these defenses deterministic was very similar. If it seems too redundant/repetitive we could address all three defense's deterministic versions at the beginning of \secref{sec:makedeterm}? We would just need to ensure that \diffpure{}'s additional seed-cycle feature is clear and seperate from the other defenses.}

\paragraph{Empirical comparison between random and deterministic versions}
\keane{We have not shown this is broken, but our experiments do seem to show that turning this to a deterministic process results in the same results as the prior work random process. We make this deterministic by setting the seed to be the same (can be in a cycle with other seeds) during the initial noising process.}
\todo{npc: this is okay}

\keane{describe the process of diffusion and de-noising}
\todo{yes do this}\keane{I had decided not to do this past what is in related work since it seemed unnecessary to understand the paper. However, I could add some details if it seems something is missing? In that case, would it be more appropriate to do so in related work rather than here?}


Similar to \randomsmoothing{}, \figref{fig:diffpure-autoattack-results} shows that \diffpure{} and its deterministic analog converge in robustness as the number of seeds used to produce subsequent noised images increases. However, the completely deterministic version of \diffpure{} (1 seed) exhibits 76\% of the robustness of the original random version ($67\%$ out of $88\%$), and the robustness is identical when sequentially alternating between only 3 separate noise patterns. This implies that the majority of the robustness from this defense comes from the deterministic de-noising process that uses the pre-trained diffusion model.

\input{figures/diffpure_autoattack_results}




%% file: figures/robust-acc-vs-num-corruptions.tex
\begin{figure}[h]
  \centering
  \begin{subfigure}{0.5\linewidth}
    \makebox[\linewidth][c]{\includegraphics[width=0.96\linewidth]{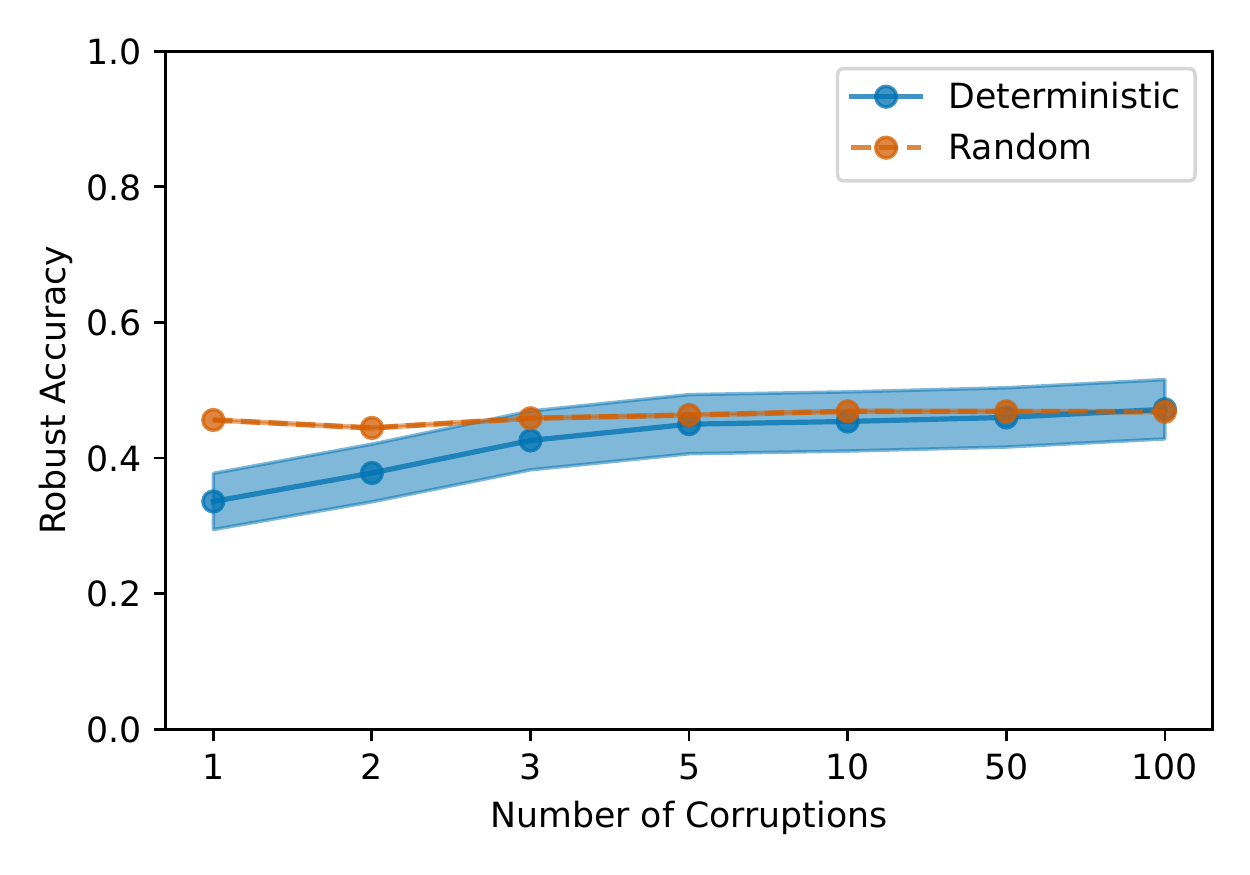}}
  \end{subfigure}

  \caption{\label{fig:robust-acc-vs-num-corruptions} From the attacker's perspective, executing PGD against randomized and deterministic smoothing converges to be the same as the number of corruptions increases. Shaded regions indicate 95\% confidence intervals.}
\end{figure}

%% file: figures/energy-based-model-converge-plot.tex
\begin{figure}[h]
  \centering
  \begin{subfigure}{0.5\linewidth}
    \makebox[\linewidth][c]{\includegraphics[width=\linewidth]{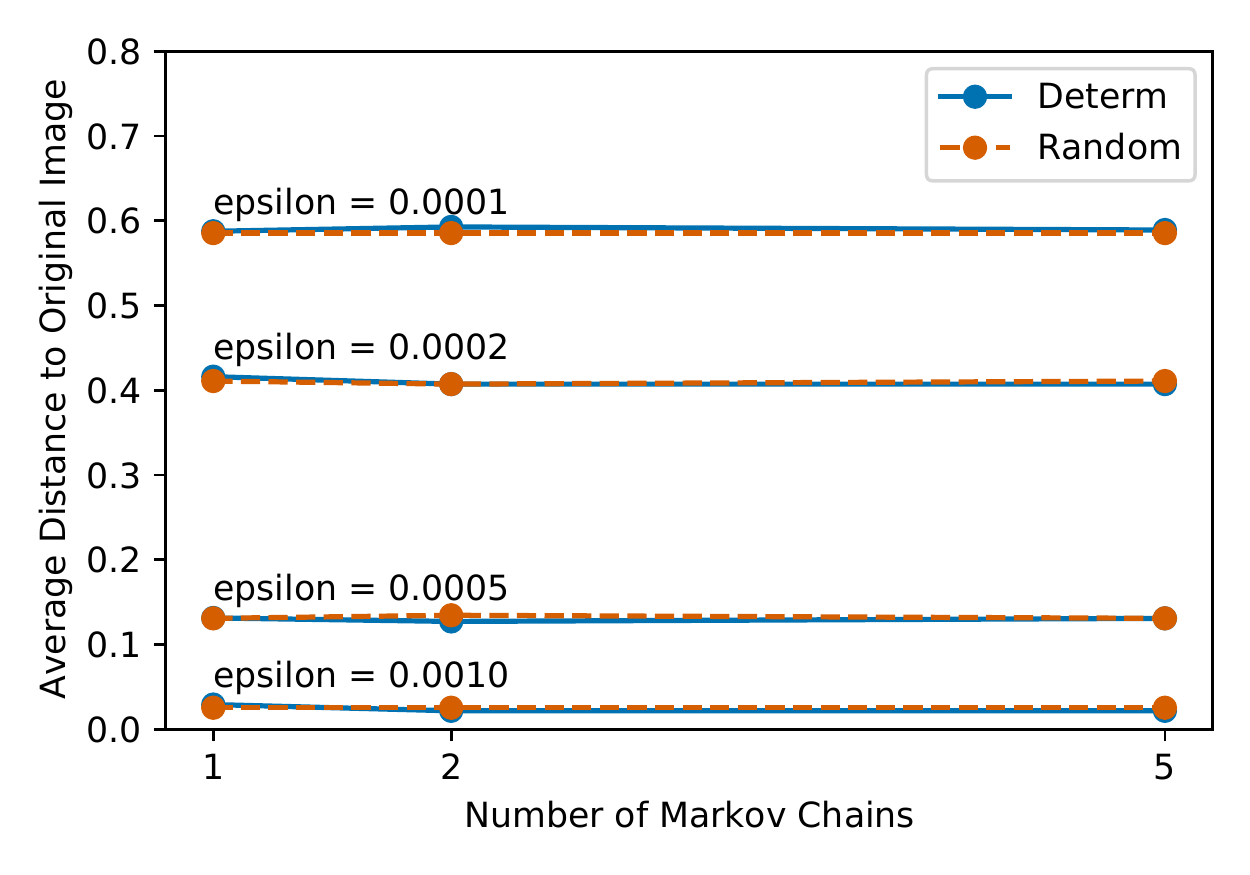}}
  \end{subfigure}
    \caption{\label{fig:ebm-converge-plot} Plot showing nearly identical robustness between attacking random and deterministic versions of \ebmdefense{} given four thresholds of adversarial example distance from original point.}
\end{figure}

%% file: figures/diffpure_autoattack_results.tex
\begin{figure}[h]
  \centering
  \begin{subfigure}{0.5\linewidth}
    \makebox[\linewidth][c]{\includegraphics[width=0.96\linewidth]{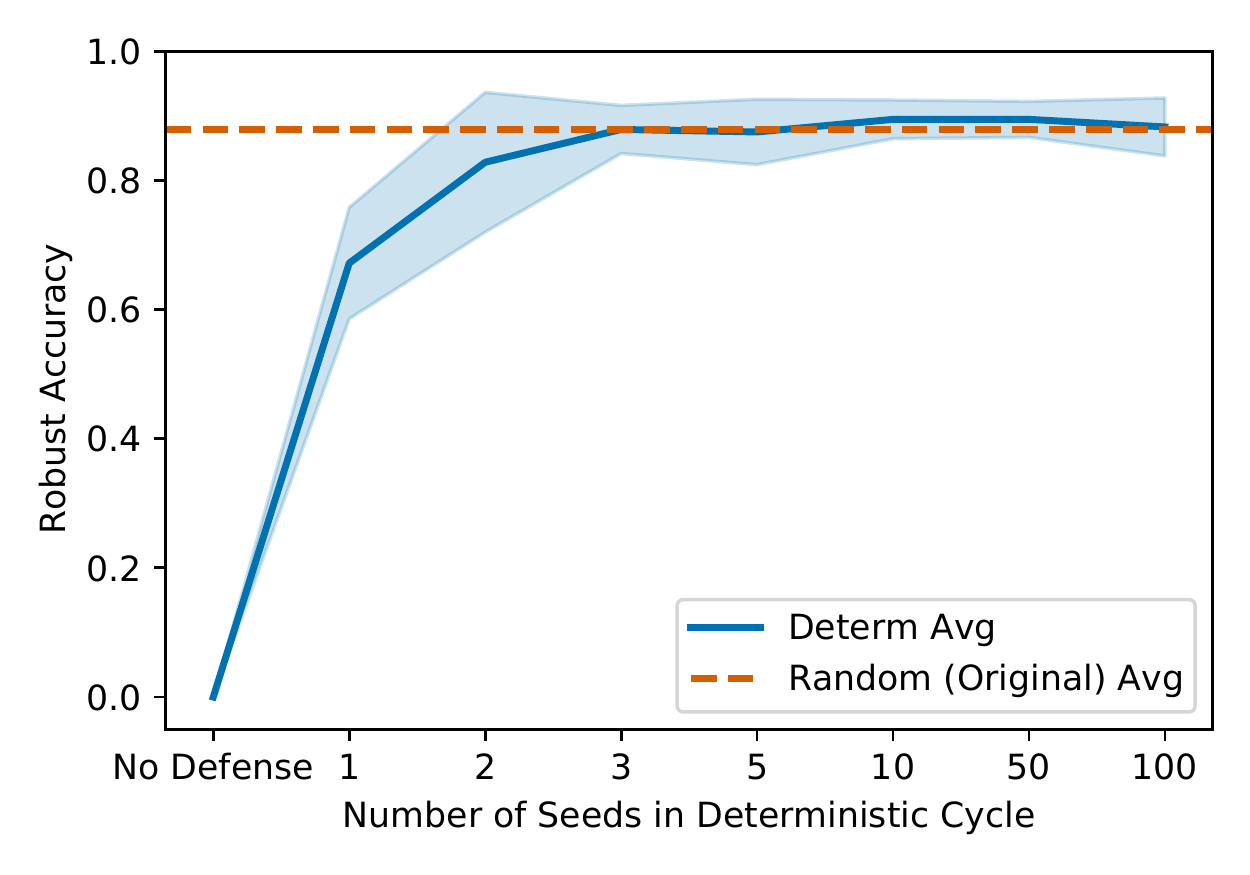}}
  \end{subfigure}
  \caption{\label{fig:diffpure-autoattack-results} A deterministic analog of \diffpure{} that uses the same noise for each inference (seed cycle length of 1) gets 67\% robust accuracy, and a version that simply alternates between two different noises when inferring (a seed cycle length of 2) gets 83\% robust accuracy, which is most of the 88\% robust accuracy the original random defense gets against AutoAttack's A-PGD.}
\end{figure}

%% file: sections/sub-space-grid-sweep.tex
\section{Subspace Grid-sweep}
\label{sec:gridsweep}

Deterministic defenses are also better because they allow a simpler brute-force robustness evaluation that we now introduce.
Our method directly searches for regions in the input space close to the original point that cause a classifier to misclassify.
We refer to these regions as \textit{adversarial regions}. By directly testing if these adversarial regions exist via a grid search, we can check to see if gradient-guided methods, used in the same space, also find these regions. If they do not, this could be indicative of obfuscated gradients~\citep{athalye2018obfuscatedgradients}.

\subsection{Searching a Subspace}
\label{sec:gridsweep:methods}

\todo{Problem: we'd like to brute force everything, but that's too slow.}\keane{wrote this paragraph and modified next paragraph to address these two comments}
Ideally, we would search the entire nearby input space for adversarial examples, but given high input dimensionality, this is not feasible.
\todo{Solution: we'll constrain the dimensionality to make it feasible.}\keane{addressed}%
We side-step this limitation by instead searching a lower-dimension (e.g., 1--6 dimensions) subspace within the nearby input space.

Subspaces are defined by $K$ orthonormal basis vectors, each of length $D$, where $D$ is the number of dimensions of the original space (e.g., $3072$ for CIFAR-10, $784$ for MNIST) and $K$ is the number of dimensions of the subspace. These vectors form a matrix $M$ with shape $K \times D$ which we use to project any point in the original space to the subspace and vice versa.

\input{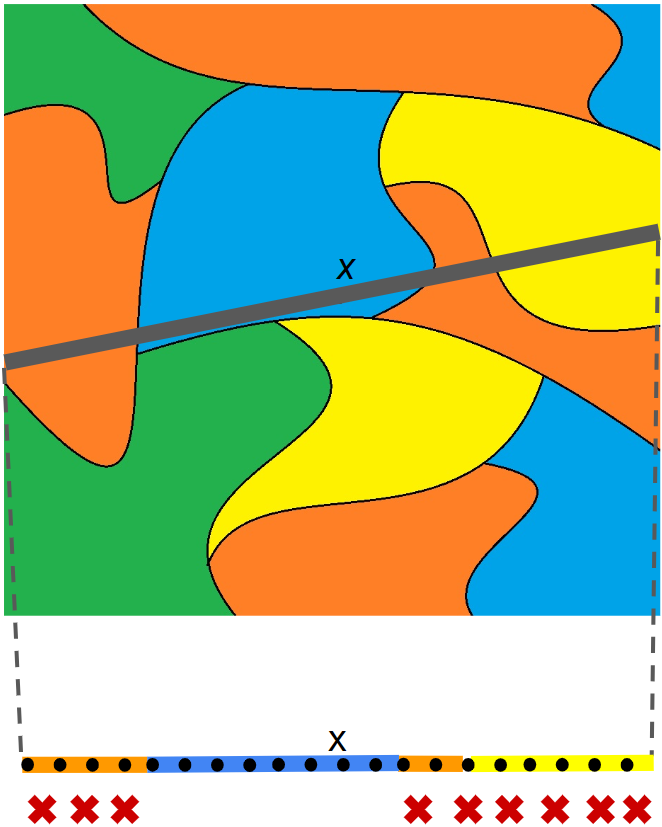}

Within this subspace, we execute a \textit{grid-sweep} search, where we divide each axis of the subspace (where each axis is defined by a basis vector) into a preset number of bins $B$ within a $K$-dimensional ball with radius $\epsilon$, where $\epsilon$ is a preset distance constraint from the original point. This makes the number of points we search in the subspace equal to $B^K$ if we use the $L_\infty$ distance, and less than or equal to $B^K$ if we use the $L_2$ distance. 

\todo{Problem: sometimes this misses regions because we have to take discrete steps.}\keane{addressed below}

\todo{Solution: we also randomly sample the same number of points but randomly.}\keane{addressed below}

However, this grid-sweep will miss adversarial regions not on the evenly spaced grid or, in higher dimensions, closer to the surface of the hyper-sphere. To alleviate this issue, we also do a random search within the subspace (and distance constraint) with the same number of points as grid-sweep. Finally, we execute PGD~\citep{Madry17AdvTraining} within the subspace to find adversarial regions. 
\keane{does this sentence make sense?} \todo{npc: Not to me, no.}\keane{I decided to comment out the sentence as it doesn't seem needed and may just confuse people.}

By evaluating with several datapoints, comparing these different methods of finding adversarial regions should be able to reveal whether gradient-guided methods are having difficulty finding adversarial regions that grid-sweep or random sampling can find. If this is the case, then it is likely the defense is not reducing the prevalence of adversarial regions, but simply making them harder to find via gradients (i.e., obfuscated gradients~\citep{athalye2018obfuscatedgradients}).


\subsection{Case study for \gridsweep{}}
\label{sec:gridsweep:results}

As shown in \secref{sec:makedeterm}, we can create a deterministic analog of published randomized defenses with similar or identical robustness as the original. 
This section gives an example of using \gridsweep{} on a deterministic published \ml{} defense, \kwta{}~\citep{Xiao2020kWTA}, that has since been shown to suffer from obfuscated gradients~\citep{Tramer2020OnAdaptiveAttacks,athalye2018obfuscatedgradients}.

\input{tables/subspace-gridsweep-summary-table}

In determining if \kwta{} is reducing the prevalence of adversarial examples within some $\epsilon$ distance of the original point or simply making them harder to find with gradient-based methods (i.e., PGD), we search subspaces consisting of 1--6 dimensions with grid-sweep, random sampling, and PGD (with 1, 10, or 20 restarts) to find examples within the subspace within $L_2$ distance of $\epsilon = 0.5$ of the original test datapoint. For grid-sweep, the number of bins varies between 9 and 1001, using fewer bins as dimensionality increases. The details of these searches and their results are in \tabref{tab:subspace-gridsweep-normal} and \tabref{tab:subspace-gridsweep-kwta} in \appref{sec:app:gridsweep}. 

We summarize the results in \tabref{tab:subspace-gridsweep-summary}. Each cell represents the fraction of vulnerable datapoints found by a specific search method compared to the union of vulnerable data-points found by all search methods. The high values present for the undefended classifier show that PGD (with 10 or 20 restarts) and Grid-sweep found nearly all vulnerable data-points that were found by all other search methods, indicating that gradient-based methods are successfully finding most adversarial regions in the undefended classifier.

\tabref{tab:subspace-gridsweep-summary} also shows the results for these same searches with a \kwta{}-defended classifier. In contrast to the undefended classifier, PGD fails to find adversarial examples on datapoints that other methods (either random sampling or grid-sweep) were able to find. This indicates that \kwta{} is not reducing the prevalence of adversarial examples, 
but is making them harder to find for gradient-based methods such as PGD.

%% file: figures/grid-sweep-visual.tex
\begin{figure}[h]
  \centering
    \includegraphics[width=\linewidth]{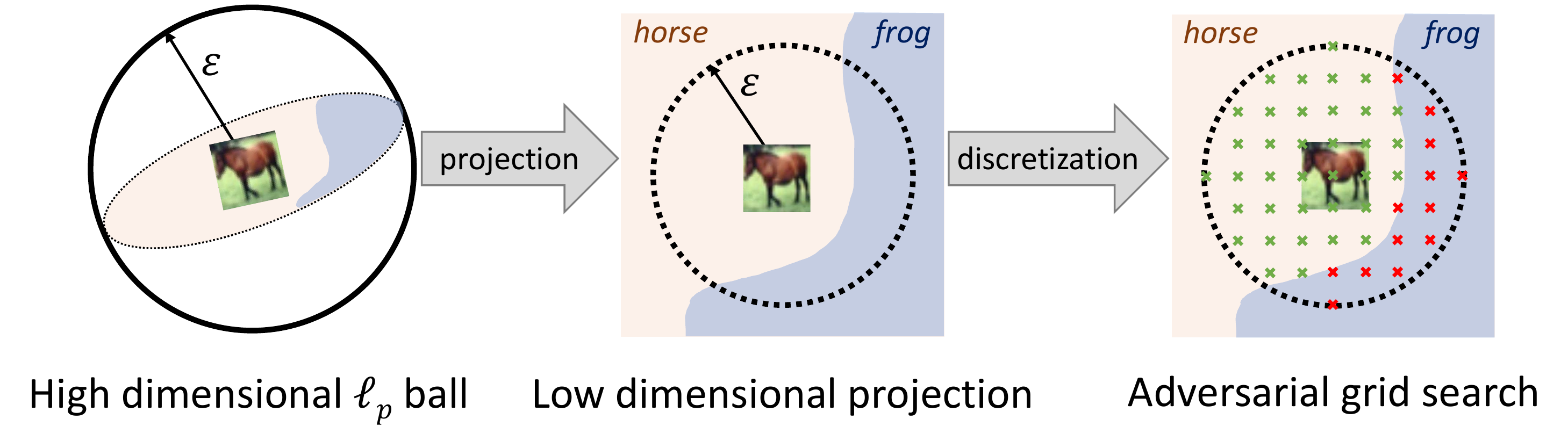}

  \caption{\label{fig:grid-sweep-visual} In order to search for adversarial regions nearby the original point, we take a small slice of the full high-dimensional space (i.e., a subspace). Within this subspace, we compare whether grid-sweep and gradient-guided attacks can find adversarial regions. If gradient-guided attacks cannot find adversarial regions that grid-sweep can, this may indicate obfuscated gradients.}
\end{figure}

%% file: tables/subspace-gridsweep-summary-table.tex
\begin{table}[!htp]\centering
\scriptsize
\begin{tabular}{@{}lrrrrrr@{}}\toprule
Model &Grid-sweep &Rand-sample &\multicolumn{3}{c}{PGD} \\\cmidrule{1-6}
& & &1 rep &10 rep &20 rep \\\midrule
Undefended &\cellcolor[HTML]{c2d2c2}0.98 &\cellcolor[HTML]{bdd9c7}0.99 &\cellcolor[HTML]{d0b5ab}0.92 &\cellcolor[HTML]{b7e1cd}1.00 &\cellcolor[HTML]{b7e1cd}1.00 \\
kWTA-defended &\cellcolor[HTML]{d1bdb3}0.95 &\cellcolor[HTML]{d0aba2}0.88 &\cellcolor[HTML]{ce6c66}0.63 &\cellcolor[HTML]{d0a69d}0.86 &\cellcolor[HTML]{d0b2a9}0.91 \\
\bottomrule
\end{tabular}
\caption{Fraction of vulnerable test datapoints (i.e. a datapoint with an adversarial example within $\epsilon=0.5$ $L_2$ distance) found by each search method compared to the union of vulnerable test datapoints found by all methods. All methods are executed on an undefended model and a \kwta{}-defended model. PGD (1, 10, and 20 repeats) finds nearly every vulnerable datapoint that Grid-sweep finds on the undefended classifier, indicating that gradient-based methods are not hindered. 
However, PGD fails to find vulnerable datapoints 
that Grid-sweep finds 
in the \kwta{}-defended classifier, indicating that this defense does not reduce the prevalence of adversarial examples, but does make them harder to find for gradient-based methods.\label{tab:subspace-gridsweep-summary}}
\end{table}

%% file: sections/discussion.tex
\section{Discussion}
\label{sec:discussion}

In this section, we 
remark on some theoretical aspects of this work and outline promising future work.

\paragraph{Problems with Randomness in Evaluation}
As pointed out in \secref{sec:relwork:defenseeval}, confidently figuring out whether an \ml{} defense can be defeated is difficult. Including randomness adds more difficulty. 

For example, randomness, sometimes by design~\citep{Xiao2020kWTA}, is sufficient to remove the ability to quickly check whether an underlying loss gradient is smooth. While this difficult loss surface could be the result of a robust defense, it also could be obfuscating a weak defense that can be defeated via an adaptive attack ~\citep{Tramer2020OnAdaptiveAttacks,Gao2022OnTLimitsStochasticDefense}. Additionally, while black-box attacks can be useful to expose gradient masking~\citep{athalye2018obfuscatedgradients}, randomness directly degrades the ability to use this class of attack for evaluation~\citep{Dong2019-benchmarkadversarialrobust}, requiring potential evaluators to forgo them as tools. Similarly, creating attacks that can expose the underlying vulnerability of randomized defenses often requires an evaluator to specially craft an attack to bypass any specific flavor of randomness included~\citep{Tramer2020OnAdaptiveAttacks}.

As shown in \secref{sec:makedeterm}, the extra effort needed to expose a randomized defense's weakness comes with little empirically verified robustness benefit. For this reason, randomized defense authors should consider providing a deterministic analog for evaluation.

\paragraph{Theoretical Aspects}
For \randomsmoothing{}, the robustness certificate gained is dependent on the Monte Carlo estimation of the distribution of classes within an $L_p$ ball of the original input point. The defense estimates this by adding Gaussian-sampled corruptions to the input point and aggregating the classifications into a distribution. However, in the version of this defense where the random seed is fixed (and known to an attacker) resulting in the same corruptions for each inference (i.e., \determsmoothing{} in \secref{sec:replay-attack:results}), a robustness certificate can no longer be justified. The loss of the robustness certificate occurs because the deterministic corruptions are no longer a valid Monte Carlo estimation\keane{is this true? Is there a better way to say this?}. In this case, an attacker no longer theoretically needs to attempt to optimize their adversarial perturbations against unknown Gaussian-sampled corruptions, but only against the known set of corruptions used in every inference.

For this reason, in principle, \determsmoothing{} should be an easier defense to attack. It is interesting, then, that the empirical attack success of an attacker that knows the deterministic corruptions used for prediction is essentially the same as an attacker that does not know the corruptions used as shown in \secref{sec:makedeterm:randomizedsmoothing} (given at least 5 corruptions are used for prediction as shown in \figref{fig:robust-acc-vs-num-corruptions}). 

\paragraph{Future Work}
Given this cursory discord between theoretical and empirical robustness, it could be valuable to explicitly investigate how an attacker's knowledge or ignorance of a fixed random seed affects their capabilities.

Also, because randomness has been shown to be effective against black-box attacks~\citep{Dong2019-benchmarkadversarialrobust}, it could be valuable to see if they become more helpful for evaluating deterministic analogs of randomized attacks. It may be the case that if a black-box attack becomes more effective against a deterministic analog of a promising randomized defense, then this may be a sign of a potentially randomness-obscured weakness in the defense.\keane{Is this true? I'm trying to talk about the idea to use black-box attacks against deterministic analogs}

%% file: sections/conclusion.tex
\section{Conclusion}
\label{sec:conclusion}

Overall, our findings suggest caution when designing or implementing \ml{} defenses that rely on randomness. Indeed, we have shown that: (1) randomness in \ml{} defenses can directly increase their vulnerability against a repetitive attacker; (2) randomness is often unnecessary to retain defense robustness; and (3) without randomness, \ml{} defenses can be evaluated with a new proposed tool, \gridsweep{}. 
We recommend that future \ml{} defenses should eschew randomness whenever possible, both in deployment and in evaluation. If randomness is included in a defense, a corresponding deterministic analog should also be evaluated.

%% file: sections/appendix.tex
\section{\gridsweep{} Details}
\label{sec:app:gridsweep}
\input{tables/subspace-gridsweep-results}

\newpage 

\section{\ebmdefense{} Details}
\label{sec:app:ebmdefense}
\input{figures/energy-based-model-dist-plot}

%% file: tables/subspace-gridsweep-results.tex

\begin{table}[ht]\centering
\scriptsize
\begin{tabular}{lrrrrrrr}\toprule
\multicolumn{2}{c}{Sub-space / Grid} & & &\multicolumn{3}{c}{PGD} \\\cmidrule{1-2}\cmidrule{5-7}
Dims &Bins &Grid-sweep &Rand-sample &1 rep &10 rep &20 rep \\\midrule
1 &1001 &\cellcolor[HTML]{b7e1cd}1 &\cellcolor[HTML]{b7e1cd}1 &\cellcolor[HTML]{d0b7ae}0.93 &\cellcolor[HTML]{b7e1cd}1 &\cellcolor[HTML]{b7e1cd}1 \\
2 &51 &\cellcolor[HTML]{ccc4b8}0.96 &\cellcolor[HTML]{b7e1cd}1 &\cellcolor[HTML]{d0a89f}0.87 &\cellcolor[HTML]{b7e1cd}1 &\cellcolor[HTML]{b7e1cd}1 \\
3 &21 &\cellcolor[HTML]{b7e1cd}1 &\cellcolor[HTML]{b7e1cd}1 &\cellcolor[HTML]{d0b2a9}0.91 &\cellcolor[HTML]{b7e1cd}1 &\cellcolor[HTML]{b7e1cd}1 \\
4 &11 &\cellcolor[HTML]{b7e1cd}1 &\cellcolor[HTML]{b7e1cd}1 &\cellcolor[HTML]{b7e1cd}1 &\cellcolor[HTML]{b7e1cd}1 &\cellcolor[HTML]{b7e1cd}1 \\
5 &9 &\cellcolor[HTML]{c7cbbd}0.97 &\cellcolor[HTML]{c7cbbd}0.97 &\cellcolor[HTML]{d0b0a7}0.9 &\cellcolor[HTML]{b7e1cd}1 &\cellcolor[HTML]{b7e1cd}1 \\
6&9 &\cellcolor[HTML]{c7cbbd}0.97 &\cellcolor[HTML]{b7e1cd}1 &\cellcolor[HTML]{d0bab0}0.94 &\cellcolor[HTML]{b7e1cd}1 &\cellcolor[HTML]{b7e1cd}1 \\
\bottomrule
\end{tabular}
\caption{Results using \gridsweep{} on an undefended model. PGD (10 repeats and 20 repeats) finds adversarial regions (within $\epsilon$ distance of original point) for every datapoint Grid-sweep finds adversarial regions, indicating that gradient-based methods are not hindered from finding existing adversarial regions.\label{tab:subspace-gridsweep-normal}}
\end{table}

\begin{table}[ht]\centering
\scriptsize
\begin{tabular}{lrrrrrrr}\toprule
\multicolumn{2}{c}{Sub-space / Grid} & & &\multicolumn{3}{c}{PGD} \\\cmidrule{1-2}\cmidrule{5-7}
Dims &Bins &Grid-sweep &Rand-sample &1 rep &10 rep &20 rep \\\midrule
1 &1001 &\cellcolor[HTML]{c7cbbd}0.97 &\cellcolor[HTML]{d0bab0}0.94 &\cellcolor[HTML]{ce645f}0.6 &\cellcolor[HTML]{d0b0a7}0.9 &\cellcolor[HTML]{ccc4b8}0.96 \\
2 &51 &\cellcolor[HTML]{d0b2a9}0.91 &\cellcolor[HTML]{d0ada4}0.89 &\cellcolor[HTML]{cf736d}0.66 &\cellcolor[HTML]{d0b2a9}0.91 &\cellcolor[HTML]{ccc4b8}0.96 \\
3 &21 &\cellcolor[HTML]{d0b7ae}0.93 &\cellcolor[HTML]{d0b0a7}0.9 &\cellcolor[HTML]{ce6c66}0.63 &\cellcolor[HTML]{d0a39b}0.85 &\cellcolor[HTML]{d0b2a9}0.91 \\
4 &11 &\cellcolor[HTML]{d0b7ae}0.93 &\cellcolor[HTML]{d0b0a7}0.9 &\cellcolor[HTML]{cf716b}0.65 &\cellcolor[HTML]{d0a39b}0.85 &\cellcolor[HTML]{d0b0a7}0.9 \\
5 &9 &\cellcolor[HTML]{ccc4b8}0.96 &\cellcolor[HTML]{d0a69d}0.86 &\cellcolor[HTML]{ce6c66}0.63 &\cellcolor[HTML]{d0a198}0.84 &\cellcolor[HTML]{d0aba2}0.88 \\
6 &9 &\cellcolor[HTML]{b7e1cd}1 &\cellcolor[HTML]{d09991}0.81 &\cellcolor[HTML]{ce6e69}0.64 &\cellcolor[HTML]{cf948c}0.79 &\cellcolor[HTML]{d09e96}0.83 \\
\bottomrule
\end{tabular}
\caption{Results using \gridsweep{} on a \kwta{}-defended model. PGD (10 repeats and 20 repeats) fails to find adversarial regions (within $\epsilon$ distance of original point) for many datapoints Grid-sweep finds adversarial regions, indicating that this defense does not reduce the prevalence of adversarial examples, but does make them harder to find for gradient-based methods.\label{tab:subspace-gridsweep-kwta}}
\end{table}

%% file: figures/energy-based-model-dist-plot.tex
\begin{figure*}[ht]
  \centering
  \begin{subfigure}{0.48\textwidth}
    \makebox[\textwidth][c]{\includegraphics[width=\textwidth]{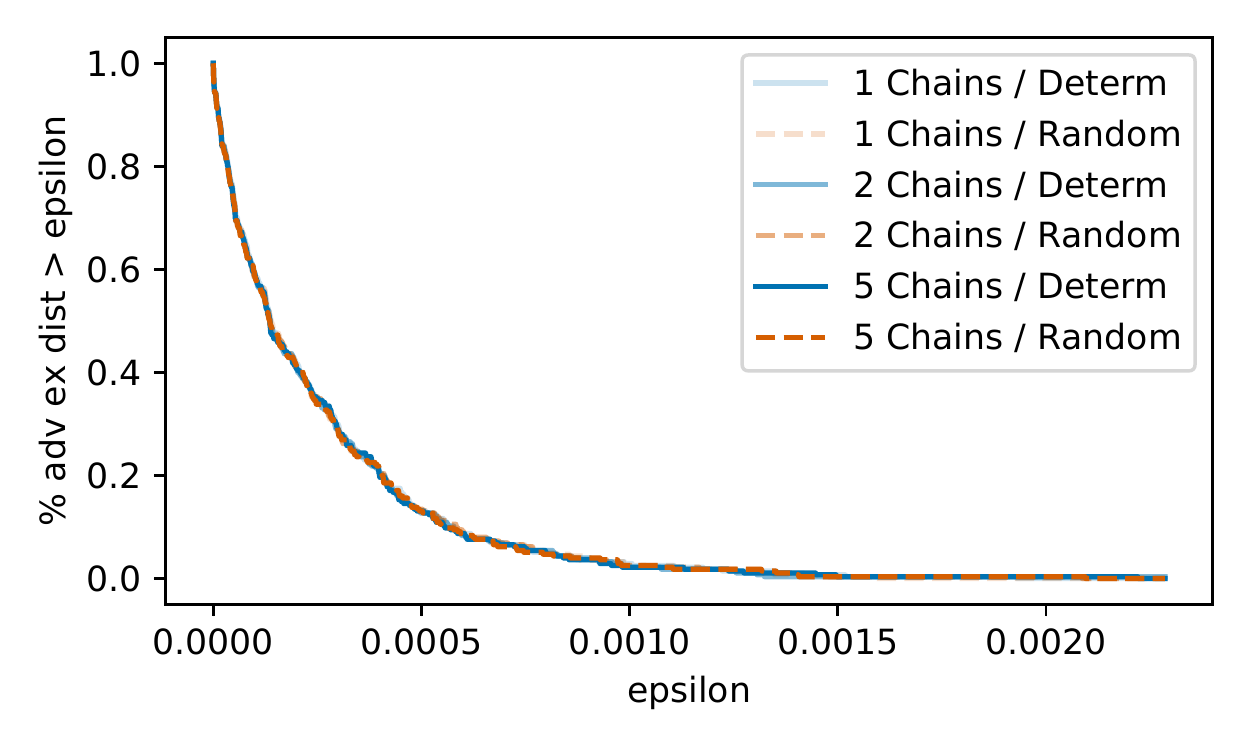}}
    \caption{\label{fig:ebm-dist-plot:full-range} The distributions of adversarial example distances found when attacking random (dashed) and deterministic (solid) \ebmdefense{} using 1, 2, or 5 Markov Chains is nearly identical.}
  \end{subfigure}
  \begin{subfigure}{0.48\textwidth}
    \makebox[\textwidth][c]{\includegraphics[width=\textwidth]{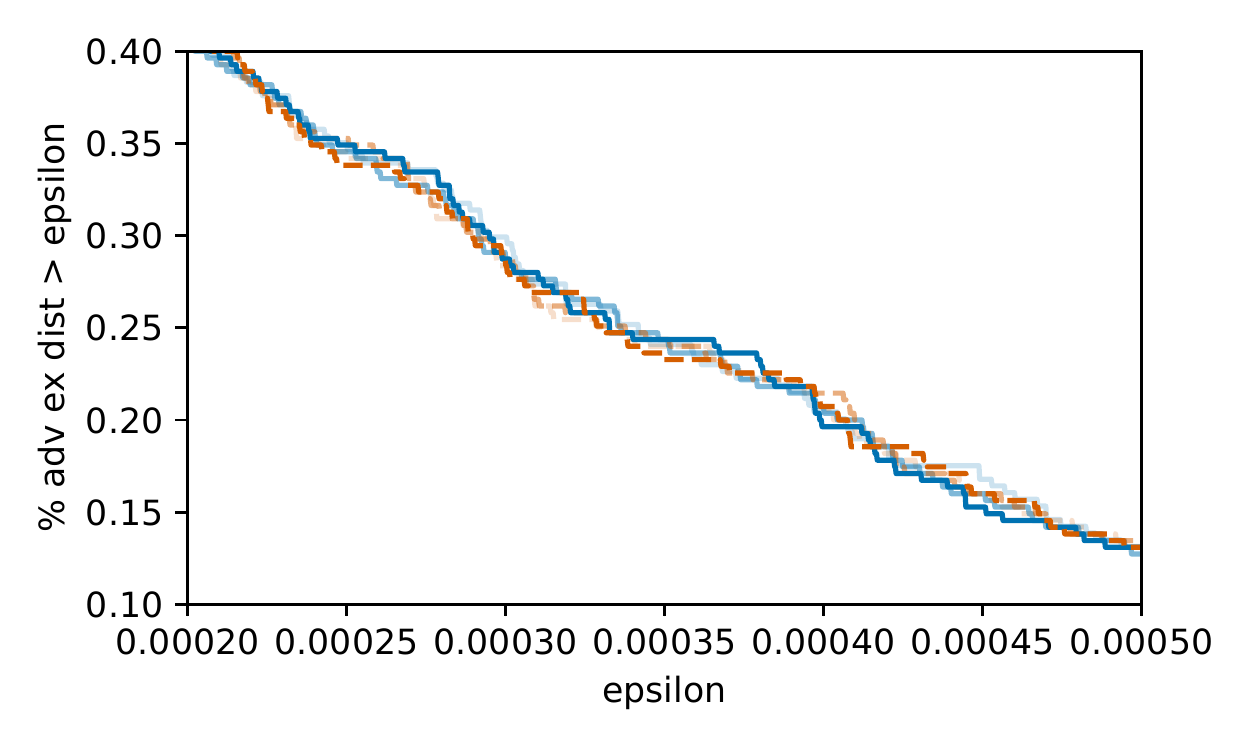}}
    \caption{\label{fig:ebm-dist-plot:zoomed-in} Zoomed in comparison of \ebmdefense{} adversarial example distances showing the slightest difference between random (dashed) and deterministic (solid) \ebmdefense{} when 1 Markov Chain is used.}
  \end{subfigure}

  \caption{\label{fig:ebm-dist-plot}Distribution of adversarial example distances from the original point when attacking a random and deterministic \ebmdefense{}.}
\end{figure*}